\documentclass{article}

\PassOptionsToPackage{numbers,sort&compress}{natbib}

\usepackage[table]{xcolor}
\usepackage{graphicx}
\usepackage{wrapfig}
\usepackage{amssymb}
\usepackage{amsmath}
\usepackage[skip=2pt]{caption}
\usepackage{array}
\usepackage{booktabs}
\usepackage{float}
\usepackage{threeparttable}
\usepackage{pifont}
\usepackage{multirow}

\usepackage[preprint]{corl_2026}

\hypersetup{
    citecolor=purple,
    pdftitle={VistaVLA: Geometry- and Semantic-Aware 3D Gaussian-Grounded VLA for Robotic Manipulation},
    pdfauthor={Mohan Liu, Zhihao Gu, Xuanyu Chen, Haitian Zhang, Kaimin Mao, Yan Wu, Wei-Yun Yau, Lin Wang}
}

\definecolor{keypurple}{HTML}{C03A9B}
\definecolor{tableBlue}{HTML}{E8F5FC}
\definecolor{lineBlue}{HTML}{8BC4E6}

\providecommand{\cmark}{}
\renewcommand{\cmark}{\textcolor{green!55!black}{\ding{51}}}

\providecommand{\xmark}{}
\renewcommand{\xmark}{\textcolor{red!70!black}{\ding{55}}}

\hypersetup{
    pdftitle={VistaVLA: Geometry- and Semantic-Aware 3D Gaussian-Grounded VLA for Robotic Manipulation},
    pdfauthor={Mohan Liu, Zhihao Gu, Xuanyu Chen, Haitian Zhang, Kaimin Mao, Yan Wu, Wei-Yun Yau, Lin Wang},
    citecolor=purple
}

\title{VistaVLA: Geometry- and Semantic-Aware 3D Gaussian-Grounded VLA for Robotic Manipulation}

\author{
  Mohan Liu$^1$,
  Zhihao Gu$^1$,
  Xuanyu Chen$^1$,
  Haitian Zhang$^1$,
  Kaimin Mao$^1$ \\
  Yan Wu$^2$,
  Wei-Yun Yau$^2$,
  Lin Wang$^{1,\dagger}$ \\
  \normalfont
  $^1$ EmPACT Lab, Nanyang Technological University, Singapore \\
  $^2$ Institute for Infocomm Research (I2R), A*STAR, Singapore \\
  $^\dagger$Corresponding Author
}
\date{}

\begin{document}
\maketitle

\begin{figure}[h!]
\vspace{-22pt}
    \centering
    \includegraphics[width=1\linewidth]{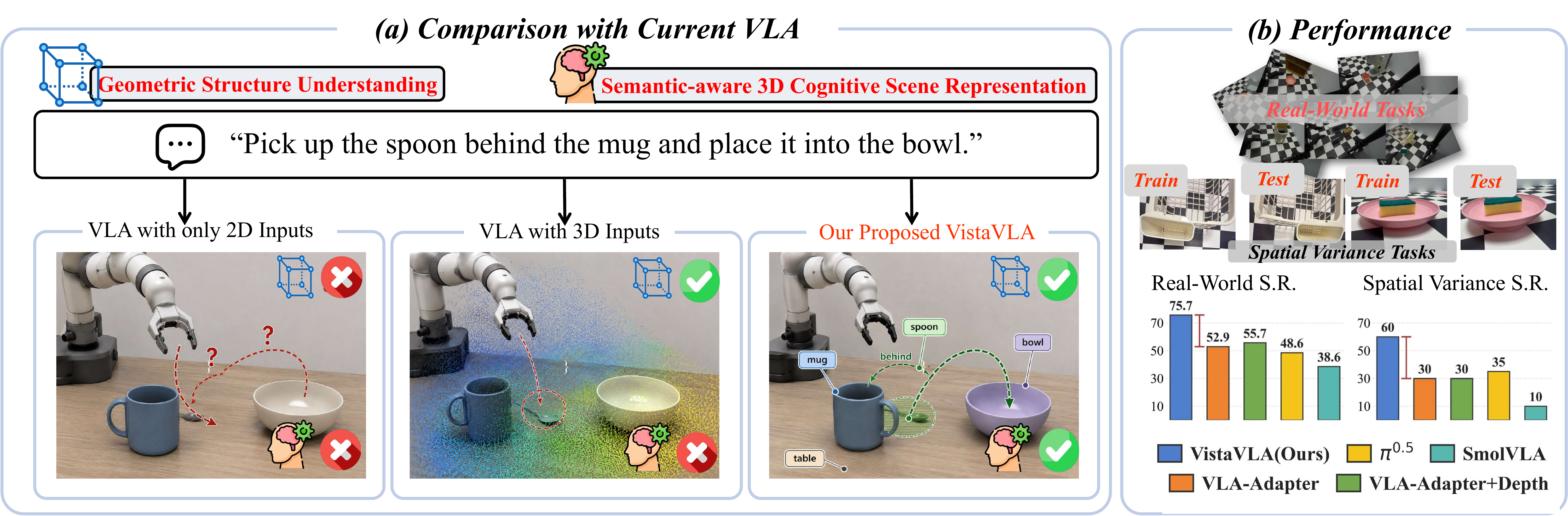}
\caption{\textbf{VistaVLA bridges 3D geometric structure and semantic grounding for VLA manipulation.}
(a) We compare current VLA paradigms: 2D-input VLAs lack explicit 3D structure, while existing 3D-input VLAs mainly provide geometric cues without sufficient semantic grounding in 3D space.
VistaVLA addresses both aspects by constructing semantically grounded 3D Gaussian scene representations for policy learning.
(b) This design improves real-world manipulation performance and robustness under spatial variations compared with strong 2D and 3D VLA baselines.}
    \label{fig:headintro}
\end{figure}

\begin{abstract}
Vision-Language-Action (VLA) models have emerged as a powerful end-to-end paradigm for robotic manipulation by mapping language instructions and 2D visual inputs directly to actions. However, these models lack an explicit, scene-level 3D representation, limiting their ability to reason over spatial layouts and geometric constraints. While recent efforts incorporate explicit 3D cues, such as depth maps or point clouds, to improve geometric awareness, they primarily capture low-level structures and lack \textit{high-level semantic grounding in 3D space}. In human cognition, interaction with the physical world relies on a \textit{3D semantic cognitive map} -- an internal mental model that integrates spatial layouts with semantic context to enable persistent, viewpoint-invariant reasoning. In light of this, we present \textbf{VistaVLA}, a novel two-stage framework that constructs a \textbf{geometry- and semantics-aware 3D cognitive representation} from 3D Gaussian primitives and grounds it as compact context tokens for VLA policy learning. Specifically, VistaVLA lifts multi-view vision-language features into 3D Gaussian primitives, forming geometry-anchored semantic tokens that align view-consistent spatial grounding with 2D visual feature spaces. To make this 3D representation computationally tractable for effective VLA control, we introduce \textbf{Merge-then-Query (MtQ)}, a token summarization mechanism. MtQ compresses dense Gaussian primitives into a highly compact set of spatially informative tokens, achieving a \textbf{99\%} token reduction while preserving action-relevant 3D layouts and semantic context. Extensive evaluations in both simulated and real-world environments demonstrate the effectiveness of VistaVLA. Notably, in real-world scenarios, VistaVLA improves success rates by \textbf{22.8\%} across seven real-world tasks and by \textbf{30.0\%} over the VLA-Adapter baseline on challenging out-of-distribution tasks.
\end{abstract}

\vspace{-10pt}
\section{Introduction}  
\vspace{-5pt}

Vision-Language-Action (VLA) models~\citep{octomodelteam2024octoopensourcegeneralistrobot,kim2024openvlaopensourcevisionlanguageactionmodel,kim2025finetuningvisionlanguageactionmodelsoptimizing,gu2026learning,black2026pi0visionlanguageactionflowmodel,intelligence2025pi05visionlanguageactionmodelopenworld,wang2025vlaadaptereffectiveparadigmtinyscale,brohan2023rt2,brohan2022rt1} have recently emerged as a powerful paradigm, shifting robot learning away from isolated task-specific policies toward generalist robotic intelligence. By training directly on large-scale multi-modal datasets, VLA models possess the unique ability to translate high-level natural language instructions and diverse visual observations directly into low-level robotic control commands. This capacity to unify open-vocabulary semantic reasoning with direct motor control makes VLA models a cornerstone for next-generation embodied AI, promising a path toward robots that can seamlessly understand and interact with complex, real-world environments. However, because these models rely primarily on 2D image observations without an explicitM spatial anchor, their ability to reason over complex geometric constraints remains severely limited, often leading to manipulation failures during precise or contact-rich actions.

Recent efforts attempt to address this bottleneck by incorporating explicit 3D cues into VLA pipelines, such as depth maps, point clouds, or sparse 3D coordinates~\citep{bhat20253dcavlaleveragingdepth,li2025qdepthvlaquantizeddepthprediction,qu2025spatialvlaexploringspatialrepresentations,li2025pointvlainjecting3dworld,li2025spatialforcingimplicitspatial}. While these methods successfully enhance basic geometric awareness, they primarily encode low-level structural geometry and fundamentally \textit{lack high-level, open-vocabulary semantic features grounded in the 3D space}, as depicted in \hyperref[fig:headintro]{\textit{\textcolor{blue}{Fig.~\ref*{fig:headintro}(a)}}}. In human cognition, physical interaction does not rely on raw, unstructured geometric data alone; instead, it depends on an internal \textit{3D semantic cognitive map} -- a mental model that unifies continuous spatial topology with functional meaning to enable persistent, viewpoint-invariant reasoning \cite{clements2006geometry, byrne1989spatial, manh2025mind, zhang2025embodied, ruan2025reactive}. Because current 3D VLA methods generate observation-centric rather than scene-level representations, they fail to bind rich multi-modal semantics to physical spatial structures. As a result, VLA policies lack a compact, action-relevant 3D context, rendering dense 3D reasoning computationally intractable for real-time robotic manipulation.

In light of this, we present \textbf{VistaVLA}, a novel two-stage framework that constructs a geometric- and semantic-aware 3D cognitive representation from Gaussian primitives \citep{kerbl2023gaussian}, grounding them into compact, policy-facing tokens for VLA policy learning, as depicted in \hyperref[fig:headintro]{\textit{\textcolor{blue}{Fig.~\ref*{fig:headintro}(a)}}}. Specifically, in \textbf{Stage I}, VistaVLA lifts open-vocabulary, language-aligned features into 3D Gaussians, yielding a scene-level representation that natively couples dense 3D geometry with high-level semantic features (Sec. \ref{sec:semantic_gaussian_field}). In \textbf{ Stage II}, these Gaussian primitives are constructed online at each control step from the latest RGB observations and calibrated camera poses, allowing the underlying 3D cognitive representation to capture real-time variations in scene layout and object configurations without relying on a persistent pre-built scene map. To ensure that reasoning over this dense 3D representation remains computationally tractable for effective control, our VistaVLA introduces a \textbf{Merge-then-Query (MtQ)} mechanism (Sec. \ref{sec:mtq}). MtQ compresses hundreds of thousands dense Gaussian primitives into a highly optimized, fixed set of compact scene-level 3D tokens. These tokens are then seamlessly integrated into the VLA via cross-attention, providing a structured, spatially anchored context for downstream action prediction.

Overall, VistaVLA achieves consistent improvements over existing 2D and 3D VLA baselines in both real-world and simulation experiments. 
In real-world manipulation, VistaVLA achieves a \textbf{22.8-point gain} over the baseline across seven tasks, as shown in \hyperref[fig:headintro]{\textit{\textcolor{blue}{Fig.~\ref*{fig:headintro}(b)}}}
Under spatial variations, VistaVLA improves depth-perturbation success from \(6/10\) to \(\mathbf{9/10}\) over the baseline. 
It also improves position-perturbation success from \(0/10\) to \(\mathbf{3/10}\), where all baselines fail. 
In simulation, VistaVLA reaches \(96.05\%\) average success on standard LIBERO and improves LIBERO-Pro-Swap from \(1.7\%\) to \(12.2\%\) over the baseline. 

In summary, our contributions are three-fold: 
(\textbf{I}) We introduce \textbf{VistaVLA}, a novel framework that mimics biological spatial cognitive awareness by bridging semantically grounded 3D Gaussians with image-based policy networks through compact, spatially anchored context tokens; 
(\textbf{II}) We present Merge-then-Query (MtQ) mechanism, an efficient token compression strategy that distills dense, unstructured Gaussian primitives from $\sim 10^5$ elements down to just 64 summary tokens—achieving a 99\% token reduction while fully retaining action-relevant 3D layouts and semantic consistency.
and (\textbf{III}) We conduct extensive real-world, simulation, and ablation studies for robotic manipulation. Results demonstrate that VistaVLA yields substantial improvements in spatial grounding, token efficiency, and robust out-of-distribution generalization capacity.

\section{Related Works}

\textbf{3D VLA Models.}
Vision-Language-Action (VLA) models achieve strong performance in 
language-conditioned manipulation~\citep{
black2026pi0visionlanguageactionflowmodel,
brohan2022rt1,
brohan2023rt2,
kim2024openvlaopensourcevisionlanguageactionmodel,
octomodelteam2024octoopensourcegeneralistrobot,
wang2025vlaadaptereffectiveparadigmtinyscale,
intelligence2025pi05visionlanguageactionmodelopenworld,
dong2026expoft,
jin2026agenticvla,
fu2026stablevla,
guo2026priorvla,
li2026rotvla,
tang2026alam,
jiang2026vlagse,
lian2026intentvla,
fan2025xr1,
koo2025hamlet,
xiao2025avavla}. 
However, these methods rely on 2D scene representation and fail to 
explicitly capture 3D scene structure and spatial relations~\citep{
kim2024openvlaopensourcevisionlanguageactionmodel,
kim2025finetuningvisionlanguageactionmodelsoptimizing,
octomodelteam2024octoopensourcegeneralistrobot,
brohan2022rt1,
brohan2023rt2,
dong2026expoft,
jin2026agenticvla,
fu2026stablevla,
guo2026priorvla,
li2026rotvla,
tang2026alam,
jiang2026vlagse,
lian2026intentvla,
li2026favla,
yang2026dyslvla,
zhong2026acotvla}.

Recent efforts to improve spatial awareness in VLAs mainly follow two directions. 
One line injects explicit geometric cues, such as point clouds, 3D coordinates, 
or depth, into the policy pipeline~\citep{
zhen20243dvla3dvisionlanguageactiongenerative,
li2025pointvlainjecting3dworld,
sun2025geovlaempowering3drepresentations,
singh2025ogvlaorthographicimagegeneration,
li2025bridgevlainputoutputalignmentefficient,
qu2025spatialvlaexploringspatialrepresentations,
bhat20253dcavlaleveragingdepth,
yuan2025depthvlaenhancingvisionlanguageactionmodels,
rao2026augvla3ddepthdrivenfeatureaugmentation,
lin2026evodepth,
song2025avi,
zhang2025falcon,
huang2025graphcotvla,
koo2025retovla}. 
Although effective, these approaches typically rely on a single 3D structural 
signal. Such signals are often poorly matched to the representations expected by 
2D-input VLM/LLM backbones and lack rich semantic 
content~\citep{
li2025bridgevlainputoutputalignmentefficient,
anatomyvla2025,
spatialrgpt2024,
li2025pointvlainjecting3dworld,
qu2025spatialvlaexploringspatialrepresentations,
sun2025geovlaempowering3drepresentations,
lin2026evodepth,
zhang2025falcon,
koo2025retovla,
huang2025graphcotvla}. 
Another line improves spatial reasoning more implicitly through 3D-aware 
alignment or spatially informed pretraining~\citep{
li2025spatialforcingimplicitspatial,
rocket2026,
zhang2025inspire,
li2025scalablevlapretraining,
ling2026gtavla,
guo2026avp,
liu2025evavla}. 
However, strong alignment may disturb the native pretrained feature 
space~\citep{
rocket2026,
zhai2023investigating,
wu2025mitigating,
li2024multimodal,
locatethenmerge2025}. 
\textit{By constrast, our VistaVLA constructs geometric- and semantic-aware 3D Gaussian primitives to establish 
3D cognitive spatial representation. 
It provides compact, multi-view-consistent, and semantic 3D scene tokens 
that can be seamlessly integrated with subsequent VLA backbones.}
\textbf{3D Gaussian Splatting (3D GS) for Robotics and Embodied Perception.}
3DGS was introduced for efficient and high-fidelity 
novel view synthesis~\citep{kerbl2023gaussian}. 
It was later extended to feed-forward 3D reconstruction from sparse 
observations~\citep{
charatan2024pixelsplat3dgaussiansplats,
chen2024mvsplat,
xu2025depthsplatconnectinggaussiansplatting}. 
Subsequent work augments Gaussian primitives with semantic or visual features. 
This enables Gaussian primitives to encode geometry and semantics in a spatially 
aligned and multi-view-consistent form~\citep{
qin2024langsplat,
zhou2024feature3dgs,
wang2024gsemsplatgeneralizablesemantic3d,
thai2025splattalk}. 
Gaussian representations have also been explored in robotics for grasping, 
world modeling, and SLAM~\citep{
zhang2026gaussiandream,
zheng2024gaussiangrasper,
ji2024graspsplats,
yu2024sparsegrasproboticgrasping3d,
lu2024manigaussian,
lu2025gwm,
chai2025gaf,
yan2024gsslam,
keetha2024splatam,
matsuki2024gaussiansplatting}.However, feature Gaussian splatting remains largely unexplored as a policy-facing representation for VLA control.
\textit{Different from these works, our method introduces the semantic gaussian primitives as 3D cognitive representation on VLA Policy control tasks.}


\noindent \textbf{Token Compression for Policy Learning.}
Efficient token processing has been widely explored in transformer-based perception and vision-language models. 
Existing methods reduce visual redundancy through adaptive token selection, dynamic pruning, or token merging~\cite{ryoo2021tokenlearner, rao2021dynamicvit, bolya2023tome}, while query-based bottlenecks summarize dense visual features into compact interfaces for language backbones~\cite{alayrac2022flamingo, li2023blip2}. 
However, these approaches mainly operate on 2D patch tokens and do not explicitly preserve 3D geometry, multi-view consistency, or scene-level spatial grounding, which are critical for robotic manipulation. 
In parallel, Gaussian compression methods improve the storage and rendering efficiency of 3D Gaussian primitives~\cite{fan2024lightgaussian, lee2024compact3dgs, wang2024contextgs}, and semantic Gaussian representations distill language-aligned or foundation-model features into 3D scenes~\cite{qin2024langsplat, zhou2024feature3dgs}. 
\textit{Different from these works, our goal is to construct compact context tokens for VLA policy learning. 
Specifically, we distill foundation 2D semantic features into 3D Gaussian primitives and further compress dense Gaussian tokens with the proposed \textbf{Merge-then-Query} mechanism, producing spatially grounded, action-relevant context tokens that can be efficiently consumed by VLA models.}

\vspace{-8pt}
\section{Methods}
\vspace{-8pt}

\begin{figure}[t]
    \centering
    \includegraphics[width=1.0\linewidth]{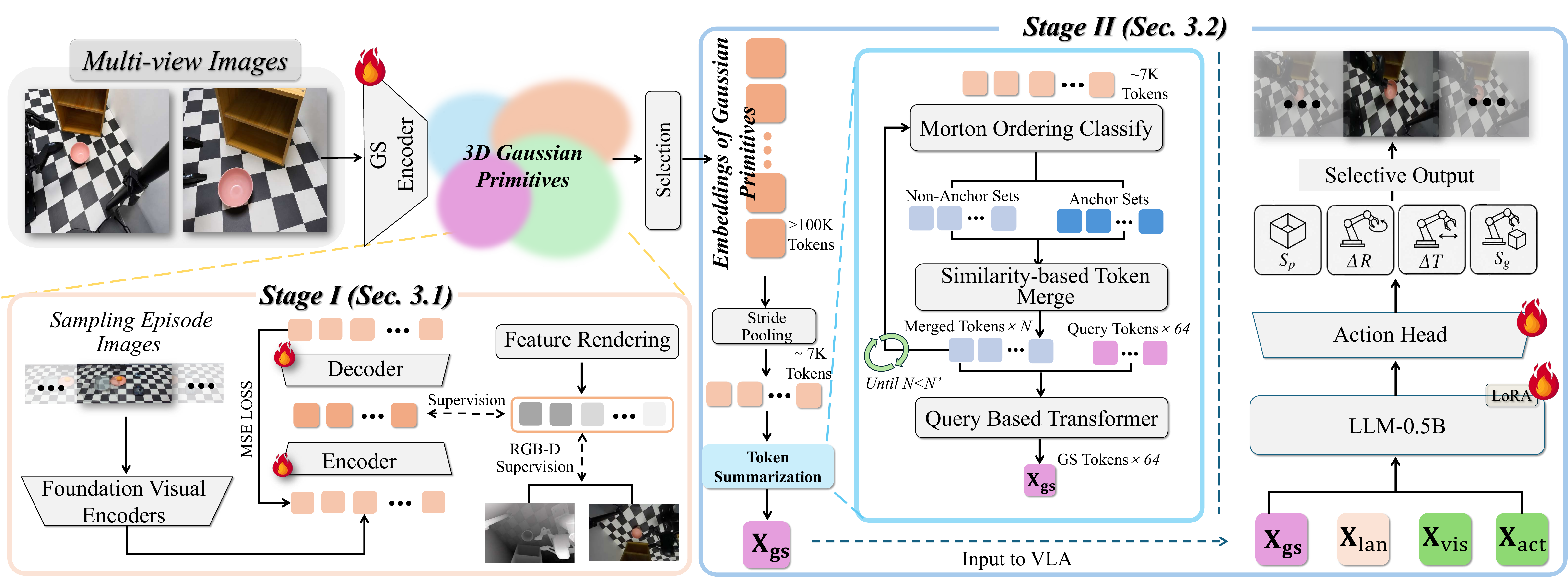}
    \caption{
    \textbf{Overview of VistaVLA.}
  Stage I: Semantic Gaussian field training distills foundation visual features under RGB-D rendering supervision.
    Stage II: Merge-then-Query compresses dense Gaussian primitives into compact 3D summary tokens.
    These tokens are finally injected into the VLA policy with instruction and wrist-image tokens for spatially grounded action prediction. 
    }
    \vspace{2mm}
    \label{fig:gsi_overview}
\end{figure}

\paragraph{Overview.}
Previous VLA models often lack explicit 3D understanding for geometry-aware spatial localization in manipulation. To this end, we aim to build a VLA framework that uses a semantic Gaussian field as a spatial context for geometry-aware manipulation, as depicted in \hyperref[fig:gsi_overview]{\textit{\textcolor{blue}{Fig.~\ref*{fig:gsi_overview}}}}. This requires addressing two key challenges: \textit{(1) how to learn Gaussian features that are both 
compact and spatially grounded, and (2) how to efficiently interface the large 
and redundant set of GS tokens with downstream VLA backbones.}
To address the first challenge, we learn a semantic Gaussian field through 
self-supervised RGB-D feature rendering and formulate its latent features as 
3D-grounded GS tokens (Sec.~\ref{sec:semantic_gaussian_field}). For the second challenge, we introduce a Merge-then-Query (MtQ) module that compresses dense GS tokens into a compact set of scene-level 3D tokens (Sec.~\ref{sec:mtq}).

\vspace{-5pt}
\subsection{Stage I: 3D Gaussian-Grounded Cognitive Representation}
\vspace{-5pt}
\label{sec:semantic_gaussian_field}

\noindent \textbf{Compact teacher feature space.}
Directly assigning high-dimensional foundation features to all Gaussian primitives is computationally expensive, as a scene typically contains a large number of primitives. We therefore first build a compact teacher feature space for Gaussian feature rendering. For each training frame, we extract dense SigLIP2~\citep{tschannen2025siglip2multilingualvisionlanguage} and DINOv2-Large~\citep{oquab2024dinov2learningrobustvisual} features, which provide complementary language-aligned semantics and robust visual descriptors. A lightweight auto-encoder is trained once to reconstruct its concatenated features from a shared bottleneck, compressing the original \(2176\)-dimensional feature into a \(128\)-dimensional latent code. This compact code is used as the teacher feature in the subsequent Gaussian feature rendering stage.

\noindent \textbf{Two-stage Gaussian feature rendering.}
Building on DepthSplat~\citep{xu2025depthsplatconnectinggaussiansplatting}, 
we represent the scene as a set of Gaussian primitives 
\(\mathcal{G}=\{g_i\}_{i=1}^{N}\). Each primitive is augmented with a learnable 
semantic feature, i.e., 
\(g_i=(\boldsymbol{\mu}_i,\boldsymbol{\Sigma}_i,\alpha_i,\mathbf{c}_i,\mathbf{z}_i)\), 
where \(\mathbf{z}_i\in\mathbb{R}^{128}\). Here, 
\(\boldsymbol{\mu}_i\), \(\boldsymbol{\Sigma}_i\), \(\alpha_i\), and 
\(\mathbf{c}_i\) denote its center, covariance, opacity, and color, respectively. Given a camera viewpoint \(v\), the renderer predicts RGB, depth, and feature 
maps as \((\hat{\mathbf{I}}^{v},\hat{\mathbf{D}}^{v},\hat{\mathbf{F}}^{v})
=\mathcal{R}(\mathcal{G},v)\), where 
\(\hat{\mathbf{F}}^{v}\in\mathbb{R}^{H\times W\times 128}\). We train the Gaussian field in two stages. \textbf{The first stage} uses 
\(\mathcal{L}_{\mathrm{stage1}}
=\lambda_{\mathrm{rgb}}\mathcal{L}_{\mathrm{rgb}} +\lambda_{\mathrm{dep}}\mathcal{L}_{\mathrm{dep}}\) to establish stable scene geometry. \textbf{The second stage} keeps the RGB-D rendering objectives and adds teacher-feature supervision, using 
\(\mathcal{L}_{\mathrm{stage2}}
=\lambda_{\mathrm{rgb}}\mathcal{L}_{\mathrm{rgb}}
+\lambda_{\mathrm{dep}}\mathcal{L}_{\mathrm{dep}}
+\lambda_{\mathrm{feat}}\mathcal{L}_{\mathrm{feat}}\), 
which distills compact semantic features into the Gaussian primitives without 
relaxing the geometric constraints. Here, 
\(\mathcal{L}_{\mathrm{rgb}}\) supervises RGB reconstruction, 
\(\mathcal{L}_{\mathrm{dep}}\) constrains the rendered and intermediate depth, 
and \(\mathcal{L}_{\mathrm{feat}}\) distills compact teacher features into Gaussian primitives.

\noindent \textbf{3D-grounded GS tokenization.}
After training the Gaussian field, we use the latent semantic feature 
attached to each Gaussian primitive as its token value. This yields the 
token set \(\mathcal{T}_{\mathrm{gs}}=\{\mathbf{z}_i\}_{i=1}^{N}\), where 
\(\mathbf{z}_i\in\mathbb{R}^{d}\) and \(d=128\). Although 
\(\mathbf{z}_i\) is represented as a compact feature vector, it is not an isolated 
2D patch descriptor; instead, it is associated with an underlying 3D Gaussian 
primitive. This grounding can be seen from the feature rendering process. For a 
target viewpoint \(v\), the rendered feature at pixel \(p\) follows the same 
alpha-compositing rule as 3D Gaussian Splatting~\citep{kerbl2023gaussian}:
{\setlength\abovedisplayskip{2pt}
\setlength\belowdisplayskip{2pt}
\begin{equation}
\hat{\mathbf{F}}^{v}(p)
=
\sum_{i\in\mathcal{N}(p)}
R_i(v,p)\mathbf{z}_i,
\qquad
R_i(v,p)=T_i^v(p)\alpha_i^v(p).
\label{eq:feature_rendering}
\end{equation}}where \(\mathcal{N}(p)\) denotes the set of projected Gaussians covering pixel \(p\).
Here, \(\alpha_i^v(p)\) is the opacity-weighted Gaussian footprint after projecting
the 3D primitive to viewpoint \(v\), and
\(T_i^v(p)=\prod_{j<i}(1-\alpha_j^v(p))\) is the accumulated transmittance over
Gaussians sorted by depth. Thus, \(R_i(v,p)\) is determined by the Gaussian
projection, opacity, depth ordering, and camera parameters. The rendered feature 
is supervised by the compact teacher feature:
{\setlength\abovedisplayskip{2pt}
\setlength\belowdisplayskip{2pt}
\begin{equation}
\mathcal{L}_{\mathrm{feat}}
=
\sum_{v\in\mathcal{V}}\sum_{p\in\Omega}
\left\|
\hat{\mathbf{F}}^{v}(p)
-
\mathbf{F}_{\mathrm{teacher}}^{v}(p)
\right\|_2^2 ,
\label{eq:feature_distillation_loss}
\end{equation}}
where \(\mathcal{V}\) denotes the set of target viewpoints and \(\Omega\) denotes 
the image pixel domain. Since this objective aggregates supervision over multiple 
viewpoints, each \(\mathbf{z}_i\) is optimized from the teacher features observed 
at all pixels where the corresponding Gaussian contributes, i.e., 
\(\{\mathbf{F}_{\mathrm{teacher}}^{v}(p)\mid R_i(v,p)>0\}\). Therefore, 
\(\mathbf{z}_i\) remains aligned with the 2D teacher feature space, while its 
supervision and correspondence are governed by the geometry and visibility of 
the underlying 3D primitive. This makes each GS token a multi-view semantic descriptor grounded in 3D space. During training, we use two calibrated RGB-D views as context and render the
third view as a novel target to enforce cross-view consistency. At inference, VistaVLA builds the 3D Gaussian field online from the policy's live
RGB observations and calibrated poses, without depth input or third view.
\vspace{-5pt}
\subsection{Stage II: Compact GS Token Summarization and Policy Injection}
\vspace{-5pt}

\label{sec:mtq}

The semantic Gaussian field typically contains \(N_0\sim 10^5\) primitives, making it impractical for downstream VLA models to process under real time constraints. To make GS tokens tractable, we introduce a Merge-then-Query (MtQ) module that converts dense Gaussian tokens into a compact fixed-length representation. MtQ first parameter-freely merges redundant GS tokens into \(N'\) tokens, and then summarizes them into \(N_q\) query tokens for downstream action prediction. 

\noindent \textbf{Parameter-free compression.}
We first reduce the dense GS token set with a parameter-free compression procedure. 
Specifically, we reshape the dense GS tokens into per-view grids and apply stride-4 average pooling to both the semantic features and their associated 3D centers, coarsely reducing the token count to roughly \(7\mathrm{K}\) before Morton-guided merging, followed by Morton-guided token merging.
Each current token maintains a semantic feature \(\mathbf{z}\), a spatial center 
\(\boldsymbol{\mu}\in\mathbb{R}^{3}\), and an aggregation size \(s\). During 
compression, \(\boldsymbol{\mu}\) is used as auxiliary geometric metadata for 
spatial partitioning, while \(\mathbf{z}\) is used for semantic similarity 
matching.

Specifically, we sort the current tokens by the Morton codes~\citep{morton1966computer} 
of their centers and split the ordered sequence alternately into an anchor set 
\(\mathcal{A}\) and a non-anchor set \(\mathcal{B}\). This produces two 
spatially distributed subsets without the iterative distance updates required by 
FPS-style partitioning. Each non-anchor token \(u\in\mathcal{B}\) is then matched 
to its most similar anchor in the semantic feature space:
{\setlength\abovedisplayskip{2pt}
\setlength\belowdisplayskip{2pt}
\begin{equation}
a^*(u)
=
\operatorname*{arg\,max}_{a\in\mathcal{A}}
\frac{\langle \mathbf{z}_u, \mathbf{z}_a\rangle}
{\|\mathbf{z}_u\|\,\|\mathbf{z}_a\|},
\qquad u\in\mathcal{B}.
\label{eq:morton_anchor_matching}
\end{equation}}We rank non-anchor tokens by their matching similarities and merge only the top 
\(50\%\) matches, avoiding forced merges between semantically dissimilar tokens. 
For each anchor \(a\), let 
\(\mathcal{M}_a=\{u\in\mathcal{B}_{\mathrm{top}}\mid a^*(u)=a\}\) denote the 
selected non-anchor tokens assigned to it. We update the anchor by size-weighted 
averaging: with the total aggregation size 
\(S_a=s_a+\sum_{u\in\mathcal{M}_a}s_u\), we set 
\(\mathbf{z}'_a=(s_a\mathbf{z}_a+\sum_{u\in\mathcal{M}_a}s_u\mathbf{z}_u)/S_a\), 
\(\boldsymbol{\mu}'_a=(s_a\boldsymbol{\mu}_a+\sum_{u\in\mathcal{M}_a}s_u\boldsymbol{\mu}_u)/S_a\), 
and \(s'_a=S_a\). Tokens not selected for merging are kept unchanged. We repeat 
this process until the token count is less than target \(N'\), where 
\(N'\ll N_0\). In our implementation, \(N'=1000\).

\begin{figure}[t]
    \centering
    \includegraphics[width=0.95\linewidth]{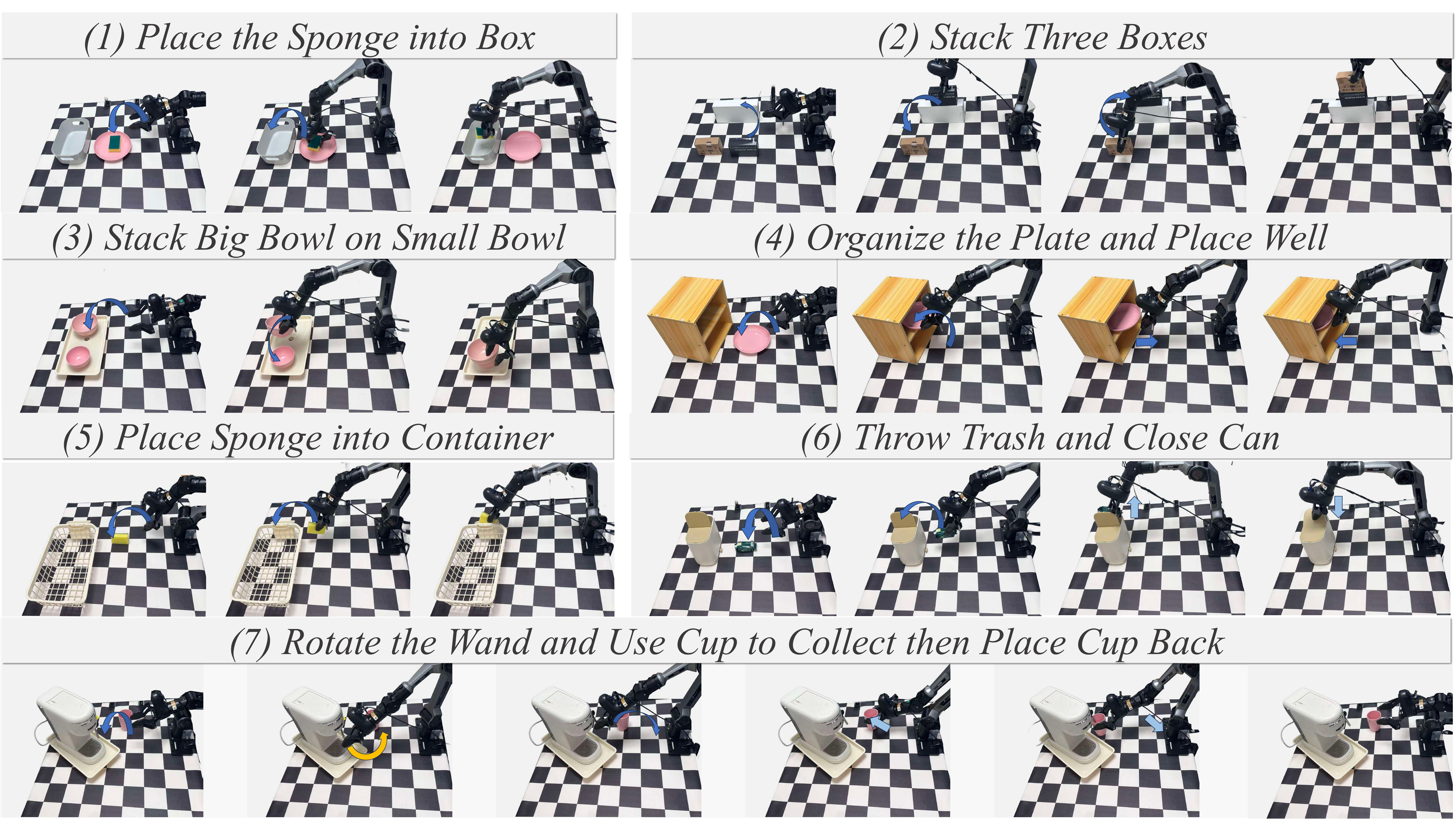}
    \caption{\textbf{Illustration of real-world tasks.} Each row shows keyframes of the corresponding manipulation trajectory, covering diverse object arrangements, interaction patterns, and temporal horizons.}\label{fig:real_world_tasks}
    \vspace{8pt}
\end{figure}

\noindent \textbf{Query-based GS summarization.}
After parameter-free compression, we further summarize the compressed GS tokens 
into a fixed-length representation for downstream VLA inference. Let 
\(\mathbf{Z}'\in\mathbb{R}^{B\times N'\times d}\) denote the compressed tokens, where \(B\) is the batch size, \(N'\) is the number of compressed tokens, and \(d\) is the feature dimension. We project \(\mathbf{Z}'\) into key and value features, and use \(N_q=64\) learnable query tokens \(\mathbf{Q}\in\mathbb{R}^{B\times N_q\times d}\) to attend to them through a two-layer Transformer decoder. The decoder output is then projected to the hidden dimension \(d_{\mathrm{llm}}\) of the downstream VLA backbone, producing 
\(\mathbf{X}_{\mathrm{gs}}=
\mathrm{Proj}_{\mathrm{llm}}(\mathrm{Dec}_{\mathrm{gs}}(\mathbf{Q},\mathbf{Z}'))\), 
where 
\(\mathbf{X}_{\mathrm{gs}}\in\mathbb{R}^{B\times N_q\times d_{\mathrm{llm}}}\). 
The resulting \(\mathbf{X}_{\mathrm{gs}}\) serves as a compact set of 
3D-grounded context tokens for downstream action prediction.

\noindent \textbf{Action generation.}
Following VLA-Adapter~\citep{wang2025vlaadaptereffectiveparadigmtinyscale}, 
we formulate continuous action prediction as causal sequence modeling, while 
injecting the GS summary tokens as additional 3D context. Specifically, we 
concatenate 2D visual tokens, language instruction tokens, GS summary tokens, 
and action query tokens as 
\(\mathbf{X}=[\mathbf{X}_{\mathrm{vis}},
\mathbf{X}_{\mathrm{lang}},
\mathbf{X}_{\mathrm{gs}},
\mathbf{X}_{\mathrm{act}}]\). 
This ordering makes the GS context language-conditioned before action decoding, 
while allowing the action query tokens to attend to all preceding visual, 
language, and 3D-grounded context tokens.
The action head predicts the next \(K\) continuous actions as 
\(\hat{\mathbf{a}}_{1:K}
=\mathcal{H}(\mathbf{H}_{\mathrm{act}},\mathbf{H}_{\mathrm{ctx}},\mathbf{s})\), 
where \(\mathbf{s}\) denotes the proprioceptive state, and 
\(\mathbf{H}_{\mathrm{act}}\) and \(\mathbf{H}_{\mathrm{ctx}}\) denote the hidden 
states of the action queries and context tokens, respectively. We train the 
policy with the regression objective 
\(\mathcal{L}_{\mathrm{act}}
=\sum_{t=1}^{K}\|\hat{\mathbf{a}}_t-\mathbf{a}_t\|_1\). 
During real-robot inference, we execute the first \(K_{\mathrm{exec}}=4\) 
predicted actions before replanning with the latest observations, reducing 
inference frequency while limiting open-loop error accumulation.

\begin{figure}
    \centering
    \includegraphics[width=1\linewidth]{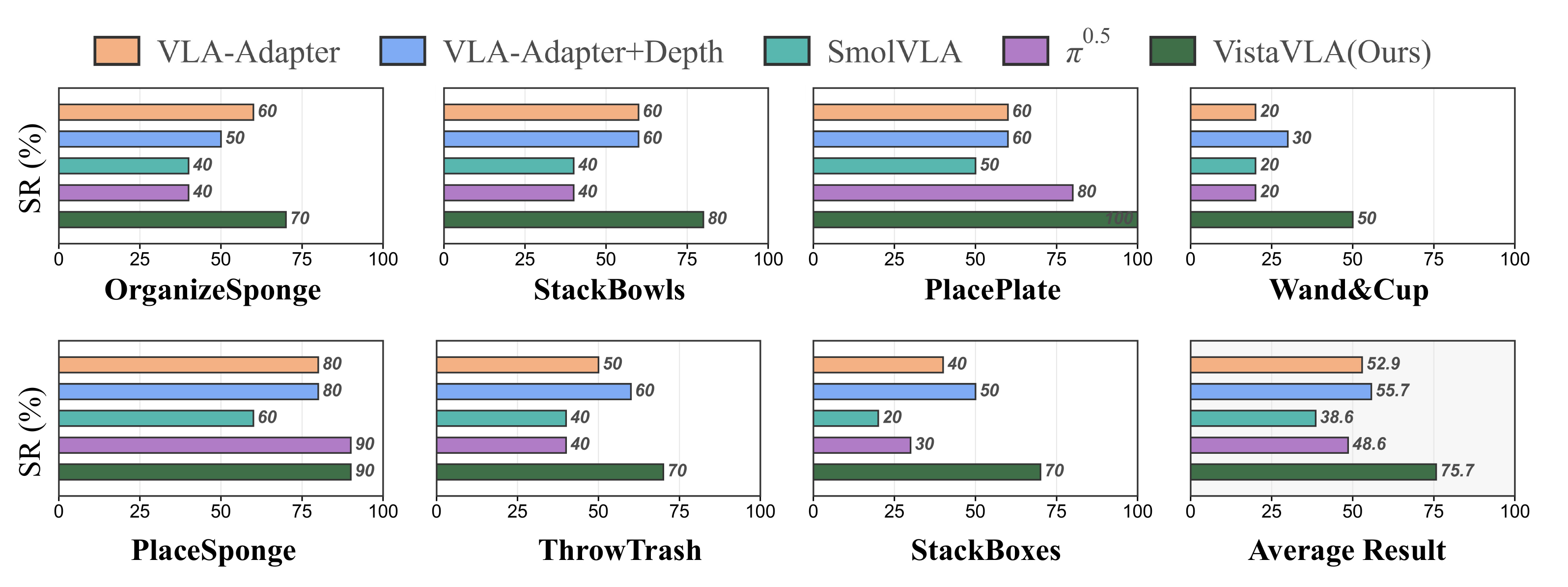}
    \vspace{-13pt}
    \caption{\textbf{Real-world results.} All policies are trained or fine-tuned on the same task-specific demonstrations. Our method consistently achieves the highest success rates across all tasks, with especially large gains on spatially demanding manipulation scenarios. Results are reported as success rates (\%).}
    \label{fig:realresult}
    \vspace{1.5em}
\end{figure}

\section{Experiments}
\vspace{-6pt}

\subsection{Real-World Experiments}
\vspace{-6pt}

\textbf{Tasks.}
We construct seven real-world manipulation tasks to evaluate the efficacy of our method. These tasks span a wide spectrum of real-world manipulation challenges, including simple pick-and-place and stacking behaviors, container organization, articulated-object interaction, and longer-horizon multi-stage operations. Concretely, the benchmark includes: 
(1)~\texttt{PlaceSponge}, 
(2)~\texttt{StackBoxes},
(3)~\texttt{StackBowls}, 
(4)~\texttt{PlacePlate},
(5)~\texttt{OrganizeSponge}, 
(6)~\texttt{ThrowTrash}, 
 and 
(7)~\texttt{Wand\&Cup}.

To further assess the robustness of the learned policy to unseen spatial configurations, we additionally introduce perturbed test variants for \texttt{PlaceSponge} and \texttt{OrganizeSponge} by modifying both the object depth and its 2D position. {{\hypersetup{linkcolor=blue}\textit{Fig.~\ref{fig:real_world_tasks}}}}.

\textbf{Setup and Baselines.}
All methods are evaluated on an \textbf{Agile-Piper manipulation platform} using 
the same task-specific demonstrations and evaluation protocol. Each method is 
tested for 10 trials per task, with the same
initial pose and the objects restored before each trial. All VLA-based methods are trained or fine-tuned under a matched effective
training budget on four RTX 5090 GPUs to ensure a fair comparison.
Our framework is built upon \texttt{VLA-Adapter}-0.5B~\citep{wang2025vlaadaptereffectiveparadigmtinyscale}, 
which also serves as our primary baseline. For a fair comparison, we equip the 
baseline with the same number of 2 fixed camera views required by our method. To provide a comparison with another 3D input modality, we further consider a depth-enhanced variant, 
\texttt{VLA-Adapter+Depth}~\citep{qu2025spatialvlaexploringspatialrepresentations}, which augments the baseline with depth 
input following the depth integration strategy of SpatialVLA~\citep{qu2025spatialvlaexploringspatialrepresentations}. We also include 
\(\pi_{0.5}\)-3B~\citep{intelligence2025pi05visionlanguageactionmodelopenworld} 
and \texttt{SmolVLA}-0.5B~\citep{shukor2025smolvlavisionlanguageactionmodelaffordable} 
as two generalist VLA baselines at different model scales, both pretrained 
on large-scale robot datasets.

\noindent \textbf{Real-World Comparison.}
As shown in \hyperref[fig:realresult]{\textit{\textcolor{blue}{Fig.~\ref*{fig:realresult}}}}, our method achieves the highest average
success rate, substantially outperforming the base \texttt{VLA-Adapter} policy
and its depth-enhanced variant. The gains are particularly pronounced on
spatially demanding tasks, where our method also surpasses the strong generalist
\(\pi_{0.5}\) policy despite using a much smaller VLA backbone. Notably,
\texttt{VLA-Adapter+Depth} provides only limited improvement over the RGB-only
baseline. We hypothesize that raw depth mainly provides observation-level
geometric cues, which are difficult to align with the semantic patch-token space
of pretrained vision backbones under limited task-specific demonstrations. In contrast, our semantic Gaussian representation grounds visual semantics in 3D space, providing a coherent, action-relevant spatial context for downstream policy learning. 
 These results suggest that the gains do not simply come from injecting
geometric cues or adding more camera views, but from introducing a structured 3D Gaussian representation as a semantic 3D spatial interface for VLA control.

\noindent \textbf{Performance under spatial variations.}
We further evaluate robustness under two task-specific spatial variation settings: height/depth variation on \texttt{PlaceSponge} and position variation on \texttt{OrganizeSponge}. As shown in \hyperref[tab:spatial_variation]{\textit{\textcolor{blue}{Table~\ref*{tab:spatial_variation}}}}, Vista-VLA achieves the best performance in both settings. 
\begin{wraptable}{r}{0.48\linewidth}
    \vspace{10pt}
    \centering
    \scriptsize
    \setlength{\tabcolsep}{1.8mm}
    \renewcommand{\arraystretch}{0.92}
    \begin{threeparttable}
    \caption{\textbf{Spatial variation robustness.}
    Depth: \textit{PlaceSponge}; Pos.: \textit{OrganizeSponge}.}
    \label{tab:spatial_variation}
    \vspace{-0.4em}
    \arrayrulecolor{lineBlue}
    \begin{tabular*}{\linewidth}{@{\extracolsep{\fill}}lcc@{}}
        \toprule[0.9pt]
        {\rlap{\color{tableBlue}\rule[-0.35em]{\linewidth}{1.45em}}\textbf{Method}} 
        & \textbf{Depth} & \textbf{Pos.} \\
        \midrule[0.4pt]
        VLA-Adapter~\citep{wang2025vlaadaptereffectiveparadigmtinyscale} & 6/10 & 0/10 \\
        +Depth~\citep{qu2025spatialvlaexploringspatialrepresentations} & 6/10 & 0/10 \\
        SmolVLA~\citep{shukor2025smolvlavisionlanguageactionmodelaffordable} & 2/10 & 0/10 \\
        $\pi_{0.5}$~\citep{intelligence2025pi05visionlanguageactionmodelopenworld} & 7/10 & 0/10 \\
        \textbf{Vista-VLA} & \textbf{9/10} & \textbf{3/10} \\
        \bottomrule[0.9pt]
    \end{tabular*}
    \arrayrulecolor{black}
    \end{threeparttable}
\end{wraptable}
On \texttt{PlaceSponge} with height/depth variation, Vista-VLA reaches 9/10 success, outperforming \texttt{VLA-Adapter}, \texttt{VLA-Adapter+Depth}, and $\pi_{0.5}$. On \texttt{OrganizeSponge} with position variation, Vista-VLA is the only method with non-zero success, improved robustness under challenging spatial shifts while also indicating that large position perturbations remain challenging.

\subsection{Simulation Experiments}
\vspace{-8pt}
\noindent \textbf{Experiment Setup.}
We conduct simulation experiments on LIBERO~\citep{liu2023liberobenchmarkingknowledgetransfer} 
as a complement to our real-world evaluation. 
We evaluate on the four standard LIBERO suites, including \textit{Spatial}, 
\textit{Object}, \textit{Goal}, and \textit{Long}. 
We further test zero-shot transfer on 
LIBERO-Pro-Swap~\citep{zhou2025liberoprorobustfairevaluation}, which introduces spatial out-of-distribution scene layouts. 
\begin{table*}[t]
    \caption{\textbf{Simulation Tasks Evaluation.}
    (a): evaluation on LIBERO\_PRO\_SWAP under spatial distribution shifts. 
    (b): representative comparison on the standard LIBERO benchmark. 
    S/O/G/L denote Spatial, Object, Goal, and Long, respectively. 
    PT denotes pretraining on large-scale robot action datasets. 
    All metrics are average success rates (\%).}
    \label{tab:libero_joint_comparison}
    \centering
    \scriptsize
    \setlength{\tabcolsep}{1.8mm}
    \setlength{\abovecaptionskip}{2pt}
    \setlength{\belowcaptionskip}{2pt}

    \begin{minipage}[t]{0.49\textwidth}
        \centering
        \vspace{0pt}
        \makebox[\linewidth][c]{\textbf{(a) LIBERO\_PRO\_SWAP}}
        \vspace{-5pt}

        {
        \renewcommand{\arraystretch}{0.92}
        \arrayrulecolor{lineBlue}
        \begin{tabular}{l c cccc c}
            \toprule[0.9pt]
            \rowcolor{tableBlue}
            Method & PT & S & O & G & L & Avg. \\
            \midrule[0.4pt]
            OpenVLA~\citep{kim2024openvlaopensourcevisionlanguageactionmodel} 
            & \cmark & 0.0 & 0.0 & 0.0 & 0.0 & 0.0 \\
            UniVLA~\citep{bu2025univlalearningacttaskcentric} 
            & \cmark & 10.0 & 0.0 & 10.0 & 0.0 & 5.0 \\
            $\pi_0$~\citep{black2026pi0visionlanguageactionflowmodel} 
            & \cmark & 0.0 & 0.0 & 0.0 & 0.0 & 0.0 \\
            Baseline~\citep{wang2025vlaadaptereffectiveparadigmtinyscale} 
            & \xmark & 6.8 & 0.0 & 0.0 & 0.0 & 1.7 \\
            \midrule[0.4pt]
            \multicolumn{7}{@{}l}{\textit{3D input methods}} \\
            \midrule[0.3pt]
            OpenVLA-SF~\citep{li2025spatialforcingimplicitspatial} 
            & \cmark & 0.0 & 0.0 & 3.6 & 0.0 & 0.9 \\
            Q-Depth VLA~\citep{li2025qdepthvlaquantizeddepthprediction} 
            & \cmark & 0.2 & 1.2 & 0.0 & 0.0 & 0.35 \\
            \textbf{Vista-VLA} 
            & \xmark & \textbf{24.0} & \textbf{10.0} & \textbf{10.0} & \textbf{4.8} & \textbf{12.2} \\
            \bottomrule[0.9pt]
        \end{tabular}
        \arrayrulecolor{black}
        }
    \end{minipage}
    \hfill
    \begin{minipage}[t]{0.49\textwidth}
        \centering
        \vspace{0pt}
        \makebox[\linewidth][c]{\textbf{(b) Standard LIBERO}}
        \vspace{-5pt}

        {
        \renewcommand{\arraystretch}{1.00}
        \arrayrulecolor{lineBlue}
        \begin{tabular}{l c cccc c}
            \toprule[0.9pt]
            \rowcolor{tableBlue}
            Method & PT & S & O & G & L & Avg. \\
            \midrule[0.4pt]
            OpenVLA~\citep{kim2024openvlaopensourcevisionlanguageactionmodel} 
            & \cmark & 84.7 & 88.4 & 79.2 & 53.7 & 76.5 \\
            $\pi_0$~\citep{black2026pi0visionlanguageactionflowmodel} 
            & \cmark & 96.8 & 98.8 & 95.8 & 85.2 & 94.2 \\
            MolmoAct~\citep{lee2025molmoact} 
            & \cmark & 87.0 & 95.4 & 87.6 & 77.2 & 86.6 \\
            GR00T N1~\citep{nvidia2025gr00tn1openfoundation} 
            & \cmark & 94.4 & 97.6 & 93.0 & 90.6 & 93.9 \\
            CoT-VLA~\citep{zhao2025cotvlavisualchainofthoughtreasoning} 
            & \cmark & 87.5 & 91.6 & 87.6 & 69.0 & 81.1 \\
            SmolVLA~\citep{shukor2025smolvlavisionlanguageactionmodelaffordable} 
            & \cmark & 93.0 & 94.0 & 91.0 & 77.0 & 88.8 \\
            \midrule[0.4pt]
            WorldVLA~\citep{cen2025worldvlaautoregressiveactionworld}
            & \xmark & 87.6 & 96.2 & 83.4 & 60.0 & 81.8 \\
            Baseline~\citep{wang2025vlaadaptereffectiveparadigmtinyscale}
            & \xmark & 91.7 & 97.2 & 96.6 & 91.0 & 94.1 \\
            \textbf{Vista-VLA} 
            & \xmark & \textbf{95.6} & \textbf{99.0} & \textbf{98.2} & \textbf{91.4} & \textbf{96.05} \\
            \bottomrule[0.9pt]
        \end{tabular}
        \arrayrulecolor{black}
        }
    \end{minipage}
\end{table*}

\noindent \textbf{Simulation Results.}
As shown in \hyperref[tab:libero_joint_comparison]{\textit{\textcolor{blue}{Table~\ref*{tab:libero_joint_comparison}}}}, Vista-VLA maintains strong 
performance on the standard LIBERO benchmark, achieving the best average result among methods without large-scale action-data pretraining. More importantly, on 
LIBERO-Pro-Swap, Vista-VLA shows better spatial OOD robustness than both 2D-input baselines and explicit 3D-input methods. These results indicate that the proposed method improves spatial generalization under constrained adaptation, while preserving competitive performance on 
standard simulation benchmarks.

\begin{table}[t]
    \centering
    \scriptsize
    \setlength{\abovecaptionskip}{2pt}
    \setlength{\belowcaptionskip}{2pt}
   \caption{\textbf{Real-world ablations.}
    (a) Camera/token ablation, where ``Cam'' counts both fixed and wrist cameras and
    1Cam is wrist-only; (b) compression ablation; (c) query-token ablation.
    T1 = PlaceSponge, T2 = StackBoxes, T3 = StackBowls, T5 = OrganizeSponge,
    and T6 = ThrowTrash.}
    \label{tab:real_world_ablations}

    \begin{minipage}[t]{0.37\linewidth}
        \vspace{0pt}
        \centering
        \makebox[\linewidth][c]{\textbf{(a) Camera/token ablation}}
        \vspace{-5pt}

        {
        \setlength{\tabcolsep}{0.75mm}
        \renewcommand{\arraystretch}{1.08}
        \arrayrulecolor{lineBlue}
        \begin{tabular}{lccccc}
            \toprule[0.8pt]
            \rowcolor{tableBlue}
            Method & Tokens & T1 & T2 & T5 & T6 \\
            \midrule[0.35pt]
            Base-1Cam & 256 & 4/10 & 2/10 & 3/10 & 1/10 \\
            Base-2Cam & 512 & 5/10 & 1/10 & 3/10 & 2/10 \\
            Base-3Cam & 768 & 8/10 & 4/10 & 6/10 & 5/10 \\
            \midrule[0.35pt]
            Vista-VLA-3Cam & 320 & \textbf{9/10} & \textbf{7/10} & \textbf{7/10} & \textbf{7/10} \\
            \bottomrule[0.8pt]
        \end{tabular}
        \arrayrulecolor{black}
        }
    \end{minipage}
    \hfill
    \begin{minipage}[t]{0.32\linewidth}
        \vspace{0pt}
        \centering
        \makebox[\linewidth][c]{\textbf{(b) Compression ablation}}
        \vspace{-5pt}

        {
        \setlength{\tabcolsep}{0.75mm}
        \renewcommand{\arraystretch}{1.08}
        \arrayrulecolor{lineBlue}
        \begin{tabular}{lccc}
            \toprule[0.8pt]
            \rowcolor{tableBlue}
            Variant & T1 & T2 & T3 \\
            \midrule[0.35pt]
            GS+FPS+Dec. & 0/10 & 1/10 & 0/10 \\
            GS+Entropy+Dec. & 4/10 & 0/10 & 0/10 \\
            GS+Merge+Linear & 7/10 & 1/10 & 0/10 \\
            \midrule[0.35pt]
            Vista-VLA (Full) & \textbf{9/10} & \textbf{7/10} & \textbf{8/10} \\
            \bottomrule[0.8pt]
        \end{tabular}
        \arrayrulecolor{black}
        }
    \end{minipage}
    \hfill
    \begin{minipage}[t]{0.20\linewidth}
        \vspace{0pt}
        \centering
        \makebox[\linewidth][c]{\textbf{(c) Query-token ablation}}
        \vspace{-5pt}
    
        {
        \setlength{\tabcolsep}{0.95mm}
        \renewcommand{\arraystretch}{0.92}
        \arrayrulecolor{lineBlue}
        \begin{tabular}{cc}
            \toprule[0.8pt]
            \rowcolor{tableBlue}
            Query Tokens & T3 \\
            \midrule[0.38pt]
            16  & 6/10 \\
            \vspace{0.15em}
            32  & 5/10 \\
            \vspace{0.15em}
            64  & 8/10 \\
            \vspace{0.15em}
            128 & 7/10 \\
            \vspace{0.15em}
            256 & 2/10 \\
            \bottomrule[0.8pt]
        \end{tabular}
        \arrayrulecolor{black}
        }
    \end{minipage}
\end{table}
\vspace{-6pt}
\subsection{Ablation Study}
\vspace{-6pt}

\noindent \textbf{Camera-view and token ablation.}
\hyperref[tab:real_world_ablations]{\textit{\textcolor{blue}{Table~\ref*{tab:real_world_ablations}(a)}}} examines whether the improvement is 
mainly caused by using more camera views or more visual tokens. We compare 
baseline variants that directly append 2D patch tokens from one, two, or three 
camera views to the downstream VLA. In contrast, Vista-VLA does not simply increase 
the number of 2D visual tokens; it converts multi-view observations into a 
compact set of 3D-grounded GS summary tokens. The results show that adding more 
2D views improves the baseline only moderately, while Vista-VLA achieves stronger 
performance with a smaller token budget, indicating that the gain comes from the 
proposed spatial-semantic interface rather than from additional image tokens.

\noindent \textbf{Compression and summarization ablation.}
\hyperref[tab:real_world_ablations]{\textit{\textcolor{blue}{Table~\ref*{tab:real_world_ablations}(b)}}} evaluates the design of the GS token 
summarization module. \textit{GS+FPS+Dec.} replaces our Morton-guided merging 
with FPS downsampling while keeping the same query decoder. 
\textit{GS+Entropy+Dec.} uses entropy-based sampling instead of our merge 
strategy. \textit{GS+Merge+Linear} keeps the proposed merge operation but removes 
the query decoder, replacing it with a linear projection. The full Vista-VLA 
performs best, suggesting that both geometry-aware token merging and query-based 
summarization are important for preserving task-relevant 3D information in a 
compact form.

\noindent \textbf{Query-token ablation.} \hyperref[tab:real_world_ablations]{\textit{\textcolor{blue}{Table~\ref*{tab:real_world_ablations}(c)}}}shows the number of query tokens on the T3. The best result is
obtained with 64 query tokens, while increasing the number of queries does not
yield consistent gains. Notably, 256 query tokens perform the worst. This
suggests that, for our two-layer Transformer network, an overly large query set
introduces redundant learnable tokens and makes the summarization bottleneck
harder to optimize, weakening the extraction of task-relevant 3D
context.

\vspace{-5pt}
\section{Conclusion, Limitation, and Future Work}
\vspace{-5pt}
\noindent \textbf{Conclusion.}
We presented VistaVLA, a two-stage VLA framework that constructs a geometry- and
semantic-aware 3D cognitive representation from Gaussian primitives and grounds
it into compact context tokens for robotic manipulation. By lifting multi-view
vision-language features into 3D Gaussian fields and summarizing dense Gaussian
tokens with Merge-then-Query, VistaVLA provides a compact, scene-level spatial
interface for VLA control. Experiments in both real-world and simulation
settings show that this semantic 3D interface improves manipulation success and
spatial generalization, especially under tasks requiring precise localization
and geometry-aware reasoning.

\noindent \textbf{Limitations and Future Work.}
Our evaluation is currently limited to tabletop manipulation with a fixed robot platform and calibrated multi-view cameras. Although this provides a controlled testbed for studying 3D-grounded VLA, broader deployment may involve larger workspace variations, dynamic occlusions, less reliable camera poses, and different robot embodiments. Also, constructing the semantic Gaussian
field still relies on posed observations.
Future work could relax these assumptions through weaker calibration requirements, manipulation-oriented feature teachers, and
evaluation in more diverse real-world environments. Nevertheless, Vista-VLA
offers a general interface for converting semantic Gaussian fields into compact
policy-facing 3D context for robotic control.


\bibliography{example}

\clearpage

\end{document}